%% file: main_iccv.tex
\documentclass[10pt,twocolumn,letterpaper]{article}
\usepackage[accsupp]{axessibility}

\usepackage{iccv}
\usepackage{times}
\usepackage{epsfig}
\usepackage{graphicx}
\usepackage{amsmath}
\usepackage{amssymb}

\usepackage{graphicx}
\usepackage{amsmath}
\usepackage{amssymb}
\usepackage{booktabs}
\usepackage{mathtools}
\usepackage{multirow}
\usepackage{titletoc}
\usepackage{tocloft}
\usepackage{appendix}

\usepackage{xcolor}
\definecolor{debianred}{rgb}{0.84, 0.04, 0.33}
\definecolor{darkgreen}{HTML}{65AE00}
\definecolor{darkergreen}{HTML}{007500}

\usepackage{pifont}
\newcommand{\cmark}{\ding{51}}%
\newcommand{\xmark}{\ding{55}}%

\newcommand{\ours}{Efficient-CLS}

\definecolor{citecolor}{HTML}{0071BC}
\definecolor{linkcolor}{HTML}{ED1C24}
\usepackage[pagebackref=true,breaklinks=true,letterpaper=true,colorlinks,bookmarks=false,citecolor=citecolor,linkcolor=linkcolor]{hyperref}

\iccvfinalcopy 

\def\iccvPaperID{9146} 

\ificcvfinal\fi

\begin{document}

\title{Label-Efficient Online Continual Object Detection in Streaming Video}

\author{Jay Zhangjie Wu$^{1}$\quad David Junhao Zhang$^{1}$\quad Wynne Hsu$^{2}$\quad Mengmi Zhang$^{3,4}$\quad Mike Zheng Shou$^1$\thanks{Corresponding Author.}
\\ \vspace{-0.3em} \\ 
$^1$Show Lab, $^2$National University of Singapore \\ 
$^3$School of Computer Science and Engineering, Nanyang Technological University, Singapore \\
$^4$CFAR and I2R, Agency for Science, Technology and Research, Singapore
}

\maketitle
\ificcvfinal\thispagestyle{empty}\fi

\begin{abstract}
Humans can watch a continuous video stream and effortlessly perform continual acquisition and transfer of new knowledge with minimal supervision yet retaining previously learnt experiences. In contrast, existing continual learning (CL) methods require fully annotated labels to effectively learn from individual frames in a video stream. Here, we examine a more realistic and challenging problem---Label-Efficient Online Continual Object Detection (LEOCOD) in streaming video. We propose a plug-and-play module, \ours{}, that can be easily inserted into and consistently improve existing CL algorithms for object detection in video streams with reduced data annotation costs and model retraining time.
We show that our method has achieved significant improvement with minimal forgetting across all supervision levels
on two challenging CL benchmarks for streaming real-world videos.
Remarkably, with only 25\% annotated video frames, our proposed method still outperforms the state-of-the-art CL models trained with 100\% annotations on all video frames. 
The data and source code will be publicly available at \url{https://github.com/showlab/Efficient-CLS}.
\end{abstract}

\section{Introduction}
\label{sec:intro}

Humans have the ability to continuously learn from an ever-changing environment, while retaining previously learnt experiences.
In contrast to human learning, prior works~\cite{aljundi2019gradient, aljundi2019online, fini2020online, wang2021wanderlust, caccia2022new} show that deep neural networks are prone to catastrophic forgetting. 
To address the forgetting problem, existing works in continual learning (CL) primarily focus on class-incremental image classification or object detection. Their experiment settings are often idealistic and simplified, where \emph{i.i.d. static images} are usually grouped by class and incrementally presented to computational models in sequence. To learn a particular task containing specific classes, an agent can go through all the data of current task \emph{over multiple epochs}. After that, the learned classes in current task become unavailable, \ie{}, \emph{no overlaps} between the sets of learned classes and unseen classes.

However, these experiment designs deviate from the online continual learning (OCL) setting in the real world, where an agent learns from \emph{temporally correlated non-i.i.d. video streams} in \emph{one single pass}. Given context regularities in natural environments, an agent is likely to encounter cases when objects of previously learnt classes \emph{co-occur} with unknown objects from unseen classes, \eg{}, a computer mouse and a computer monitor often co-occur. Taking these considerations, \cite{wang2021wanderlust} introduces OCL on object detection in real-world video streams. They evaluate existing CL approaches on this setting and report a huge performance gap compared with offline training. 

Based on the setting in \cite{wang2021wanderlust}, we take a significant step further and introduce a novel problem setting called Label-Efficient Online Continual Object Detection (LEOCOD), which highlights two unique challenges. \textbf{First}, the setting in \cite{wang2021wanderlust} is such that the CL algorithms are trained with every mini-batch over multiple passes. We tighten the training recipe in LEOCOD to strictly online, where data is allowed to have one single pass and models are trained on the entire video dataset for only one epoch. \textbf{Second}, existing CL models require fully supervised training where box-level ground truth labels of every object on every video frame have to be obtained from human annotators. Unlike static images, acquiring human annotations for object detection on videos can be expensive and daunting. Thus, in LEOCOD, the video frames per mini-batch are sparsely annotated to alleviate the burdens of extensive human labeling, making LEOCOD one step closer to real world application.

\begin{figure*}[th!]
\centering
  \includegraphics[width=0.98\linewidth]{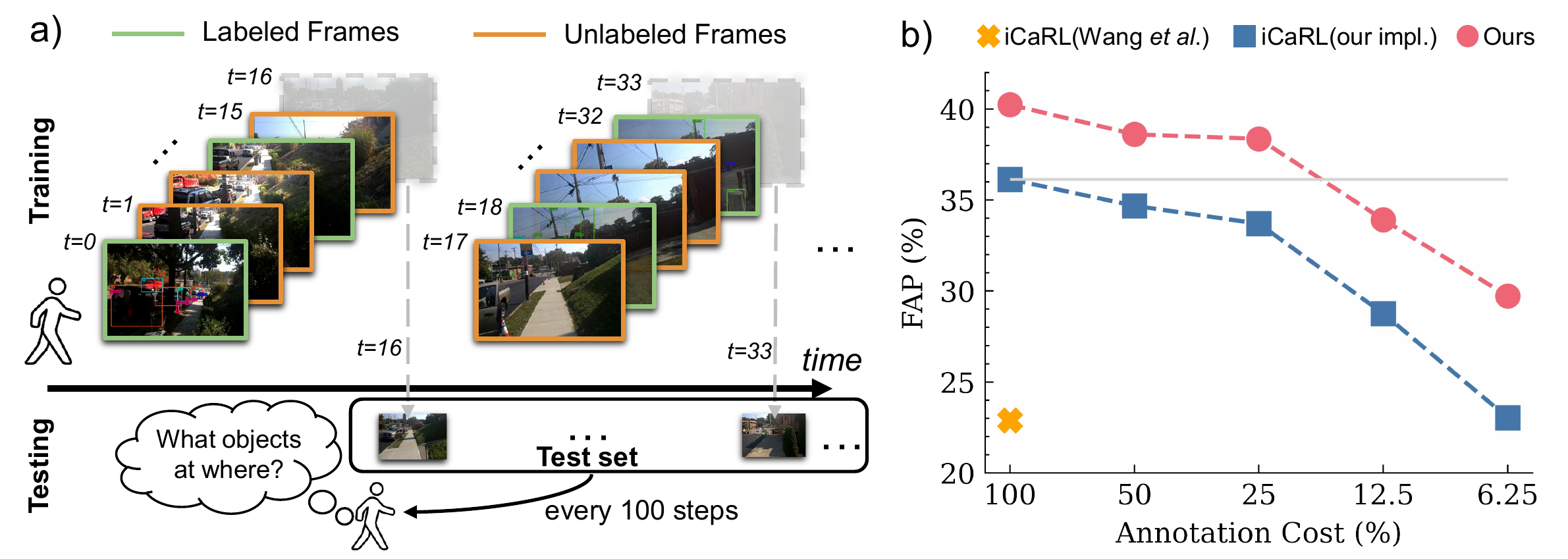}
  \caption{
  \textbf{(a) Problem introduction}: An agent continuously learns from a never-ending online video stream over time. In each training step, out of a mini-batch containing 16 consecutive video frames, only a fixed proportion of frames are labeled (green boundary), while the rest of the frames are unlabeled (orange boundary). Following~\cite{wang2021wanderlust}, the video frame after every training mini-batch (transparent) is held out for testing. After every 100 training steps, the agent is evaluated on all the video frames from the test set for object detection.
  \textbf{(b) Key results}: Our proposed method (red) consistently outperforms the best competitive baseline (blue) by a margin of 5\%. Remarkably, our model, trained at 25\% annotation cost, surpasses the best baseline trained at 100\% (grey line). The orange cross denotes the performance of the state-of-the-art model, which is 15\% lower than our method.
  }
  \vspace{-0.8em}
  \label{fig:teaser}
\end{figure*}

Cognitive science works~\cite{wang2020generalizing, lake2017building} show that humans are efficient at continuously learning from very few annotated data samples. We get inspirations from the theory of Complementary Learning Systems (CLS) in human brains~\cite{kumaran2016learning}, and propose a plug-and-play module for the LEOCOD task, dubbed as \ours{}. 
In \ours{}, we introduce two feed-forward neural networks as slow and fast learners. In the fast learner, memory is rapidly adapted to the current task. The weights of the slow learner change a little on each reinstatement, and are maintained by taking the exponential moving average (EMA) of the fast learner's weights over time.  
Though a few continual learning models in previous works~\cite{arani2022learning,pham2020contextual} also use a similar source of inspiration, they miss the effect of reciprocal connections from slow learners to fast learners, which we intend to address. 
Inspired by the bidirectional interaction in CLS~\cite{ji2007coordinated}, we reactivate the weights of the slow learners to predict meaningful pseudo labels from the unlabeled video frames and use these pseudo labels to guide the training of the fast learner, closing the loop between the two systems. 

We demonstrate the \emph{versatility} and \emph{effectiveness} of our \ours{} on two standard real-world video datasets, OAK~\cite{wang2021wanderlust} and EgoObjects~\cite{egoobjects}.
Our proposed method can be easily integrated into existing CL models and consistently improve their performance by a large margin in LEOCOD. It is worth noting that, with only 25\% labeled data, our method surpasses the comparative baselines trained with full supervision (Figure~\ref{fig:teaser}(b)). 

To summarize, we make the following key contributions:
\begin{itemize}
    \itemsep0em 
    \item We introduce a new, challenging and important problem of label-efficient online continual object detection (LEOCOD) in video streams. Solving this problem would greatly benefit real-world applications in reducing annotation cost and model retraining time.
    \item We propose \ours, a plug-and-play module inspired from the Complementary Learning Systems (CLS) theory, which can be integrated into existing CL models and learn efficiently and effectively with less supervision and minimal forgetting.
    \item We benchmark existing CL methods on the task of LEOCOD and demonstrate the state-of-the-art performance of our method through extensive experiments.
\end{itemize}

\section{Related Work}

\subsection{Continual Learning} 
To alleviate catastrophic forgetting, many continual learning methods exploit an external buffer where a limited number of old samples are stored and used for replay when adapting to a new task. iCaRL~\cite{rebuffi2017icarl} stores the representative exemplars in past tasks for knowledge distillation and prototype rehearsal. Gradient Episodic Memory (GEM)~\cite{lopez2017gradient} formulates optimization constraints on the exemplars in memory. Averaged GEM (A-GEM)~\cite{chaudhry2018efficient} is an improved version of GEM that achieves faster training and less memory consumption. GDumb~\cite{prabhu2020gdumb} greedily stores samples in memory as they come and trains a model using samples only in the memory. Dark Experience Replay++ (DER++)~\cite{buzzega2020dark} combines replay with knowledge distillation and regularization, and samples logits along the entire optimization trajectory.

In contrast to classical continual learning where data are separated by task boundaries and models are trained with multiple iterations in every task, we examine a more realistic and challenging problem where data are provided in tiny batches and models are trained on these batches only once. Recently, online continual learning (OCL) has gained increasing interests in computer vision~\cite{aljundi2019gradient, caccia2022new, shim2021online, chen2020mitigating, wang2021wanderlust}. Many OCL methods rely on representative memory replays to prevent forgetting. \cite{aljundi2019gradient} utilizes gradients of network parameters to select replay samples of maximum diversity. Subsequent works~\cite{aljundi2019online,shim2021online} propose to use losses and scoring functions as criteria for selecting the most representative samples for replay. However, these approaches tackle image classification problem in an artificial setting, where new classes appear in a specific order. Their performance in real-world vision tasks remains unclear. 

Lately, \cite{wang2021wanderlust} benchmarks CL methods in the real-world online setting with full supervision. As the video streams arrive endlessly in a real-time manner, assigning annotations to all the video frames for training computational models is laborious and time-consuming. It becomes even more daunting in object detection tasks where class labels and bounding boxes of all objects on a video frame have to be provided. Reducing burdensome costs of labeling remains an under-explored and challenging problem in online continual object detection. We propose a self-sustaining \ours{}, which is capable of exploiting the unlabeled video frames by pseudo-labeling when the number of labeled frame is limited.

\subsection{Complementary Learning Systems} 
The essence of fast and slow learning in Complementary Learning Systems (CLS) has benefited several continual learning algorithms in image recognition~\cite{pham2020contextual,pham2021dualnet,rostami2019complementary,arani2022learning,kamra2017deep}. However, these methods either require the task boundaries, which are not applicable in our online video setting, or they require to train fast and slow learning systems with replay samples from the same replay buffer, which could easily lead to overfitting problem when the replay buffer has limited capacity. To eliminate overfitting problem, \cite{rostami2019complementary} and \cite{kamra2017deep} utilize generative replay models to couple sequential tasks in a latent embedding space. While generative approaches have succeeded in artificial and simple datasets, they often fail in complex vision tasks, \eg{}, object detection. 
DualNet~\cite{pham2021dualnet} also employs a slow-fast learning architecture. However, its update process for the slow learner is more computationally demanding because it leans on both self-supervised and supervised learning objectives. The self-supervised phase demands additional training iterations on extensive batches to yield satisfactory results, thereby delaying the training of the fast learner. In contrast, our method updates the slow learner in real-time using the fast learner's weights, obviating the need for distinct slow learner training. Our approach is optimized for online CL scenarios where computational resources are constrained.

\subsection{Semi-Supervised Learning}
The goal of semi-supervised learning (SSL)~\cite{tarvainen2017mean, berthelot2019mixmatch, yang2022survey, yang2022class} is to reduce the demand of labeled data and harness unlabeled data for performance improvement. \cite{smith2021memory} first introduce SSL to the context of CL, and propose a strategy that combines pseudo-labeling, consistency regularization, out-of-distribution detection, and knowledge distillation to solve the problem of class-incremental image classification. However, to distill knowledge from previous tasks, their model relies heavily on the task boundary that identifies the change of training classes, which is not available in our LEOCOD setting. To reduce annotation costs in object detection, several methods~\cite{jeong2019consistency, sohn2020simple, liu2021unbiased} capitalize the teacher-student networks. In general, a teacher model predicts pseudo labels or enforces a consistency loss to guide the student networks.
However, these previous works on semi-supervised object detection only consider the offline setting on static image datasets, while none of them has been extended to online continual learning on dynamic video streams. 
We discovered new insights that pseudo-labeling from slow learner to fast learner can not only reduce annotation overhead, but also alleviate forgetting, which is critical to the design of efficient continual learners in real world.

\begin{figure*}[t!]
\centering
  \includegraphics[width=0.96\textwidth]{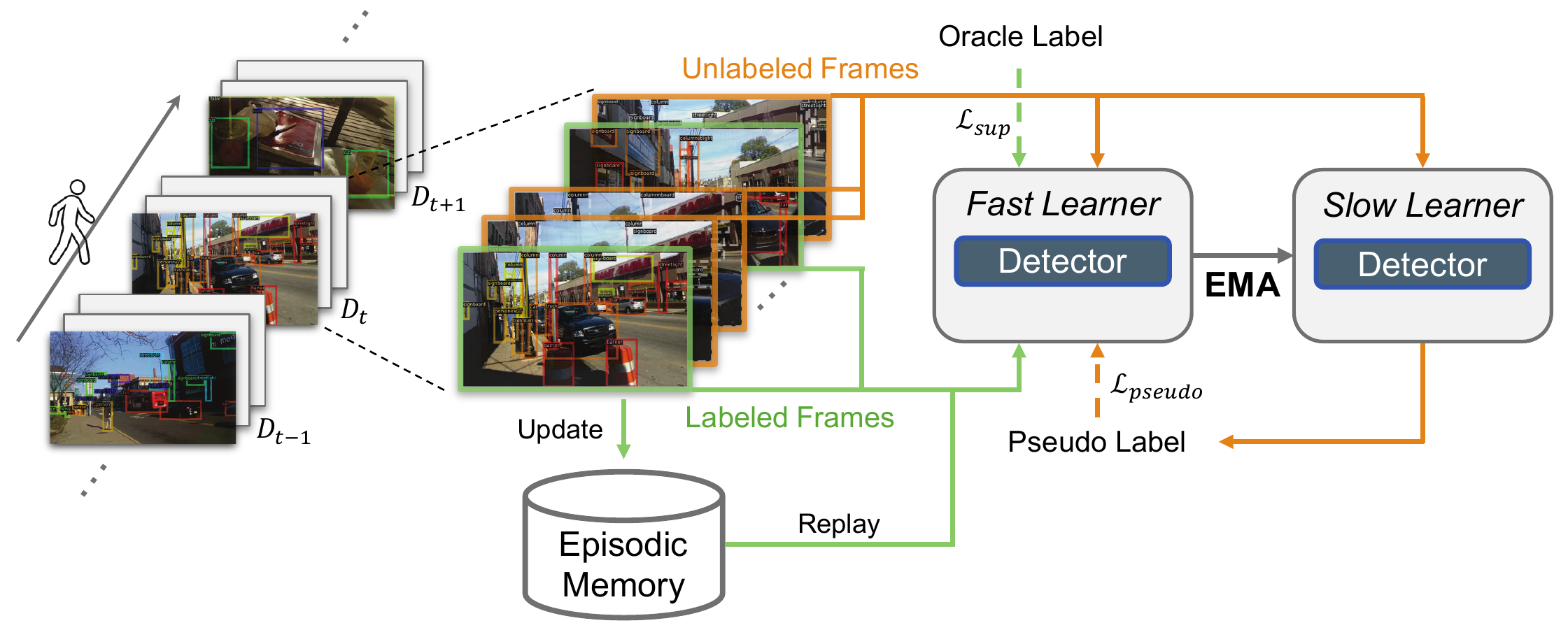}
  \caption{
  \textbf{The framework of \ours{}}. At each learning step, the system receives a batch of temporally continuous data $D_t$, including labeled (green) and unlabeled (orange) frames. The fast learner trains the labeled frames alongside a small subset of labeled exemplars retrieved from episodic memory with the supervised loss $\mathcal{L}_{sup}$. Meanwhile, the fast learner leverages the pseudo labels generated by the slow learner to optimize a pseudo loss $\mathcal{L}_{pseudo}$. To reinstate memory of the slow learner, the weights of the slow learner are updated by taking the Exponential Moving Average (EMA) of the fast learner’s weights. The fast and slow learners are complementary to each other, forming a positive feedback loop. 
  }
  \vspace{-0.8em}
  \label{fig:framework}
\end{figure*}

\section{Method}

\subsection{Problem Setting}

We advance the online continual object detection in \cite{wang2021wanderlust} to a label-efficient and computationally-efficient setting---Label-Efficient Online Continual Object Detection (LEOCOD). In contrast to the setting in \cite{wang2021wanderlust} where video frames are \emph{extensively annotated} and trained with \emph{multiple epochs}, an agent in LEOCOD continuously learns from a \emph{sparsely annotated} video stream in \emph{a single pass} over time (see Figure~\ref{fig:teaser}(a)). 

Formally, we consider the online continual object detection on a continuum of video streams $\mathcal{D}=\{D_1, \cdots, D_T\}$ where at time step $t$, a learning agent receives a mini-batch of continuous video frames $D_t$ from current environment for online training (one single pass). To perform label-efficient object detection, within the batch $D_t$, only a subset of video frames $D_t^{s}=(X_t^{s}, Y_t^{s})$ are labeled, while the remaining video frames $D_t^{u}=(X_t^{u})$ are unlabeled. For each labeled data sample, its annotation contains the bounding box locations and their corresponding class labels. 

\subsection{Efficient Complementary Learning Systems}\label{sec:method}

We propose a plug-and-play module dubbed as \ours{}. Specifically, it consists of two feed-forward networks: (i) the fast learner is designed to quickly encode new knowledge from current data stream and then consolidate it to the slow learner; and (ii) the slow learner accumulates the acquired knowledge from fast learner over time and guides the fast learner with meaningful pseudo labels, when full supervision is not available. Same as \cite{rebuffi2017icarl, chaudhry2018efficient, prabhu2020gdumb, buzzega2020dark, wang2021wanderlust}, we maintain an external episodic memory, as a replay buffer, to store exemplars that can be retrieved for replays alongside ongoing video stream. As the fast and slow learners are model agnostic, our \ours{} can be easily integrated into existing CL models, which leads to less supervision and minimal forgetting.

\vspace{-0.5em}
\paragraph{Learning with Labeled Frames.}
The fast learner and slow learner use the same standard Faster-RCNN \cite{ren2015faster} detector $f$. Despite the same architecture, the weights of the fast and slow learners are not shared. We use $\theta_{F}$ and $\theta_{S}$ to denote the network parameters for fast and slow learners respectively. As shown in Figure~\ref{fig:framework}, at each training step $t$, we use the labeled video frames $D_t^{s}=(X_t^{s}, Y_t^{s})$ to optimize the fast learner $\theta_{F}$ with the standard supervised loss $\mathcal{L}_{sup}$ in Faster-RCNN \cite{ren2015faster}. It consists of four losses: Region Proposal Network (RPN) classification loss $\mathcal{L}_{cls}^{rpn}$, RPN regression loss $\mathcal{L}_{reg}^{rpn}$, Region of Interest (ROI) classification loss $\mathcal{L}_{cls}^{roi}$, and ROI regression loss $\mathcal{L}_{reg}^{roi}$. We define $\mathcal{L}_{sup}$ as:
\begin{equation}
    \begin{multlined}
    \mathcal{L}_{sup} = 
    \mathcal{L}_{cls}^{rpn}(X_t^{s}, Y_t^{s}) + \mathcal{L}_{reg}^{rpn}(X_t^{s}, Y_t^{s}) + \\
    \mathcal{L}_{cls}^{roi}(X_t^{s}, Y_t^{s}) + \mathcal{L}_{reg}^{roi}(X_t^{s}, Y_t^{s}). 
    \end{multlined}
\end{equation}

\vspace{-0.5em}
\paragraph{Learning with Unlabeled Frames.}
We introduce a pseudo-labeling paradigm to capitalize the information from unlabeled video frames $D_t^{u}=(X_t^{u})$ for training. In our early exploration, we intuitively use the fast learner for pseudo-labeling as it quickly adapts the knowledge of nearby frames. However, we observe that using the pseudo labels generated by the fast learner for self-replay exhibits biases towards recently seen objects, which is less effective in preventing forgetting. This has also been verified in our ablation study (Section~\ref{sec:ablation}). In contrast, the slow learner preserves the semantic knowledge over a longer time span which generates pseudo labels with fewer biases. This encourages the fast learner to capture more generic scene representations, hence, in turn, contributing to reinstatement of memory in the slow learner, resulting in a positive feedback loop. 

Given all these design considerations, the slow learner takes the unlabeled video frames $D_t^{u}$ as inputs to estimate the possible objects of interest and their corresponding bounding box locations. For brevity, we refer these ``pseudo bounding boxes and their corresponding class labels" as ``pseudo labels" in the paper. To get rid of false positives, we apply a threshold $\tau$ to filter out bounding boxes with predicted low confidence scores. Moreover, there also exist repetitive boxes which negatively impact the quality of pseudo-labeling. To address this issue, we use the technique of class-wise non-maximum suppression (NMS) \cite{ren2015faster} to remove the overlapped boxes and get the high-quality pseudo labels. Formally, the procedure of pseudo label generation is summarized below:
\begin{equation}
    Y_t^{u} = \mathrm{NMS}([f(X_t^{u}; \theta_{S})]_{>\tau}), 
\end{equation}
where $[\cdot]_{>\tau}$ denotes the bounding box selection with confidence score larger than $\tau$. 

Given that the video streams are captured from the egocentric perspective in the real world, head and body motions may lead to undesired motion blur effects on some video frames. To enforce our module to learn invariant object representations from these video frames, same as the previous work \cite{zoph2020learning}, we apply data augmentation techniques on the pseudo-labeled frames, including 2D image crops, rotations, and flipping. Note that different from image classification, the predicted bounding box locations also need to be updated accordingly after image augmentations. We denote these pseudo-labeled video frames and their re-adjusted pseudo labels after data augmentations as ($\tilde{X}_t^{u},\tilde{Y}_t^{u}$). We can then use these pseudo pairs ($\tilde{X}_t^{u},\tilde{Y}_t^{u}$) to train the fast learner by optimizing the pseudo loss $\mathcal{L}_{pseudo}:= \mathcal{L}_{cls}^{roi}(\tilde{X}_t^{u}, \tilde{Y}_t^{u}) + \mathcal{L}_{reg}^{roi}(\tilde{X}_t^{u}, \tilde{Y}_t^{u})$. 

Overall, our \ours{} is jointly trained with the following losses:
$\mathcal{L}_{total} = \mathcal{L}_{sup} + \lambda_{pseudo}\mathcal{L}_{pseudo}$, where $\lambda_{pseudo}$ is the weight of $\mathcal{L}_{pseudo}$.

\vspace{-0.5em}
\paragraph{Synapses Consolidation via EMA.}
To alleviate forgetting of obtained knowledge, we apply Exponential Moving Average (EMA) to gradually update the slow learner with the fast learner's weights. The evolving weight changes in the slow learner are functionally correlated with the memory consolidation mechanism in the hippocampus and the neocortex \cite{arani2022learning}. Formally, we define EMA process as:
\begin{equation}\label{equ:EMA}
    \theta_{S} = \alpha \theta_{S} + (1-\alpha) \theta_{F},
\end{equation}
where the $\alpha \in [0,1]$ is EMA rate. According to the stability-plasticity dilemma, a smaller $\alpha$ means faster adaption but less memorization. Empirically, we set $\alpha = 0.99$, which leads to best performance. 


\section{Experimental details}\label{sec:exp}

\subsection{Datasets}
We evaluate the continual learning methods on two realistic and challenging video datasets,  OAK~\cite{wang2021wanderlust} and EgoObjects~\cite{egoobjects}, for the task of online continual object detection. 

\textbf{OAK}~\cite{wang2021wanderlust} is a large egocentric video stream dataset spanning nine months of a graduate student’s life, consisting of 7.6 million frames of 460 video clips with a total length of 70.2 hours~\cite{singh2016krishnacam}. The dataset contains 103 object categories. We follow~\cite{wang2021wanderlust} in the ordering of training and testing data splits. One frame every 16 consecutive video frames lasting for 30 seconds is held out to construct a test set and the remaining frames are used for training. 

\textbf{EgoObjects}~\cite{egoobjects} is one of the largest object-centric datasets focusing on object detection task. 
The dataset contains around 100k video frames with $\sim$250k annotations, which correspond to 277 categories and 1110 main objects~\cite{pellegrini20223rd}. Additionally, the dataset follows a long-tailed distribution which makes the task more challenging and real-world oriented. 
We adopt the same rule as OAK dataset to split the training and testing set.

\subsection{Baselines}\label{sec:baseline}

We compare our model against the following baselines: 1) \textbf{Vanilla training}: \emph{Incremental Training} is a naive baseline trained sequentially over the entire video stream without any measures to avoid catastrophic forgetting; \emph{Offline Training} is an upper bound that can access all the data throughout training; 2) \textbf{Continual learning algorithms}: \emph{EWC} \cite{kirkpatrick2017overcoming}, \emph{iCaRL} \cite{rebuffi2017icarl}, \emph{A-GEM} \cite{chaudhry2018efficient}, \emph{GDumb} \cite{prabhu2020gdumb}, \emph{DER++} \cite{buzzega2020dark} and \emph{iOD} \cite{kj2021incremental}.

The iCaRL model implemented by \cite{wang2021wanderlust} stands as the SOTA method in online continual object detection. We reproduce their results using the released code\footnote{https://github.com/oakdata/benchmark}. When calculating RPN and ROI losses for replay samples, their iCaRL model neglects the losses of background proposals and penalizes the foreground losses according to the proportion of the current samples and replay samples. We empirically found that this trick hinders the model from effective episodic replay, thus resulting in severe forgetting. Therefore, we re-implement the iCaRL by discarding the re-weighting trick and reverting back to the standard RPN and ROI losses. We name these two different implementations as \emph{iCaRL(Wang \etal{})} and \emph{iCaRL(our impl.)}, respectively.

\input{tables/main_results}

\subsection{Evaluation}\label{sec:eval}

\paragraph{Protocols.}
First, we define the annotation cost as the proportion of number of labeled frames versus the total 16 frames within a mini-batch $D_t$. For example, if 2 out of 16 consecutive frames within $D_t$ get labeled, the annotation cost is $2/16=12.5\%$. The frames to be labeled are randomly selected within each mini-batch $D_t$. Considering that different choices of labeled frames might influence the model performance, for fair comparisons between models, we fix the choice of randomly selected labeled frames and use the same labeled and unlabeled frames for training all models. 
Based on the various annotation costs, we introduce two training protocols: \emph{fully supervised protocol} (100\% annotation cost) and \emph{sparse annotation protocol} (where the annotation cost is less than 100\%). In sparse annotation protocol, we further split the training experiments based on 50\%/25\%/12.5\%/6.25\% annotation costs.

\paragraph{Metrics.}

We evaluate the models with three standard metrics: continual average precision (CAP), final average precision (FAP) and forgetfulness (F) \cite{wang2021wanderlust}. We employ an Average Precision (AP) metric at an Intersection over Union (IoU) threshold of 0.5, commonly referred to as AP50.

\textbf{CAP} shows the average performance over the entire video stream. That is,  
\begin{equation}
    {\rm CAP} = \frac{1}{N}\sum_{i=0}^{N} {\rm CAP}_{t_i} = \frac{1}{NC}\sum_{i=0}^{N} \sum_{c=0}^{C} {\rm CAP}_{t_i}^{c},
\end{equation}
where ${\rm CAP}_{t_i}^{c}$ is the AP of class $c$ on test set at the $i$-th evaluation step, $N$ is the total number of evaluation steps.

\textbf{FAP} is the final performance of a model after seeing the entire video. That is, ${\rm FAP} = {\rm CAP}_{t_N}$, where $t_N$ denotes the last evaluation step. 

\textbf{F} estimates the forgetfulness of the model due to the sequential training. It takes into account the time interval between the presence of an object category and its subsequent presence. For a class $c$, we sort the ${\rm CAP}_{t_i}^c$ according to the time interval $k$ between evaluation time $t_i$ and the last time $t_i - k$ the model is trained on $c$. After ${\rm CAP}_{t_i}^c$ is sorted, all ${\rm CAP}_{t_i}^c (i=0,\cdots,T)$ are divided into $K$ bins $B_{kmin}, \cdots, B_{kmax}$ according to the time interval $k$. The average ${\rm CAP}$ (${\rm aCAP}_{k}$) of each bin $B_k$ is defined as the model's performance for detecting class $c$ after the model has not been trained on $c$ for $k$ time steps. The forgetfulness (F) of the class $c$ is defined as the weighted sum of the performance decrease at each time:
\begin{equation}
    {\rm F}^c = \sum_{k=kmin}^{kmax} \frac{k-kmin}{\sum_{k=kmin}^{kmax} k-kmin} \times ({\rm aCAP}_{kmin} - {\rm aCAP}_{k}).
\end{equation}
The overall forgetfulness is: ${\rm F} = \frac{1}{C} \sum_{c=0}^{C} {\rm F}^c$.

\subsection{Implementation Details}

For a fair comparison, we followed the prior work \cite{wang2021wanderlust} to use Faster-RCNN \cite{ren2015faster} with ResNet-50 backbone \cite{he2016deep} as our object detection network, which is initialized by the weights pre-trained on PASCAL VOC \cite{everingham2015pascal}. We used Adam optimizer with a constant learning rate of 0.0001, and the batch size was set to 16 frames. Same as \cite{wang2021wanderlust}, we maintained a replay buffer with 5 samples per class. At each time step $t$, we first randomly retrieved 16 video frames from the replay buffer for joint training. We used confidence thresh $\tau=0.7$ to generate pseudo-labels for unlabeled frames, and applied EMA rate $\alpha=0.99$ to update the slow learner. The weight of pseudo loss $\lambda_{pseudo}$ was set to 1.0. All the experiments were conducted on 2 NVIDIA RTX 3090 GPUs. 


\begin{figure*}[h!]
\centering
  \includegraphics[width=0.98\textwidth]{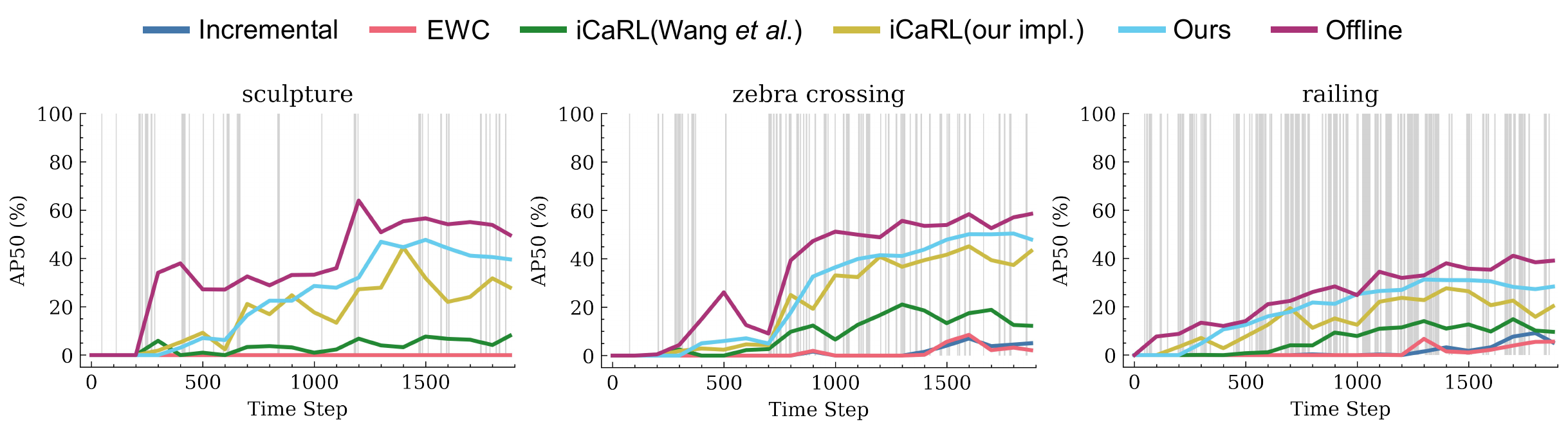}
  \caption{\textbf{The changes of ${\rm CAP}_{t_i}^{c}$ with sampled categories on OAK dataset.} The x-axis denotes time step across the entire video stream. The y-axis denotes the AP50 of the category at specific time step (\ie{}, ${\rm CAP}_{t_i}^{c}$). The grey line indicates the existence of the category. }
  \label{sfig:time_ap}
  \vspace{-0.3em}
\end{figure*}

\section{Results}

\subsection{Fully Supervised Protocol}
As the previous work \cite{wang2021wanderlust} focuses on online continual object detection (OCOD) in video streams, we first evaluated model performance in fully supervised setting (\ie{}, 100\% annotation cost), where all video frames are paired with human labels. The reported results were measured by standard metrics (CAP, FAP, and F, Section~\ref{sec:eval}) in Table~\ref{tab:main}. 

\vspace{-0.5em}
\paragraph{Performance of CL baselines.}
\cite{wang2021wanderlust} benchmarked Incremental, EWC, iCaRL(Wang \etal{}), and Offline Training on OAK dataset. They found the replay-based method (\ie{} iCaRL(Wang \etal{})) outperforms regularization-based method (\ie{} EWC) by 10\% in FAP, while iCaRL(Wang \etal{}) has a huge gap of 30\% compared with Offline Training. Similar observations were made in Table~\ref{tab:main}, but the performance of Incremental and EWC were 4\% lower than that in \cite{wang2021wanderlust}, as in our setting we only trained each mini-batch of video frames once (they trained each mini-batch 10 times). 
As mentioned in Section~\ref{sec:baseline}, we introduce several variations to the original design of iCaRL(Wang \etal{}). Compared with iCaRL(Wang \etal{}), we observed a huge performance boost in FAP from 22.89\% to 36.14\% on OAK and from 37.61\% to 60.80\% on EgoObjects, abridging the gap between the baseline and Offline Training. 

We further adapted other standard CL baselines, including iOD, A-GEM, GDumb, DER++, to the OCOD setting for comparisons. iOD is the SOTA method in offline class-incremental object detection. Though it performs well in prior setting, iOD collapses when adapted to online video streams. One explanation is that iOD requires explicit task boundary to trigger the reshape of model gradients that optimizes knowledge sharing between adjacent tasks. However, in online video streams, the task boundaries by classes are no longer available and the change of tasks is hard to identify, resulting in the failure to prevent forgetting. In contrast, replay-based methods (\ie{}, A-GEM, GDumb and DER++) generalize much better to the real-world streaming video. 

\vspace{-0.5em}
\paragraph{Performance w/ \ours{}.}
Inheriting from the benefit of fast and slow learning with EMA, our \ours{} consistently improves all the state-of-the-art CL methods (\ie{} iCaRL(our impl.), A-GEM, GDumb and DER++) by a significant margin. Taking iCaRL(our impl.) for example, with \ours{}, we observed superior performance and minimal forgetting compared to baseline methods, even when categories appeared infrequently (\eg{}, \emph{sculpture} in Figure~\ref{sfig:time_ap}). Since semantic contextual information is more important in indoor environments on EgoObjects compared to the outdoor environments in OAK, we noticed that the improvement brought by our method is even greater on EgoObjects with an increase of 6.25\% in FAP, 3.95\% in CAP \% and 3.07\% in F. Consistent performance gains are also noticeable for A-GEM, GDumb and DER++, demonstrating the effectiveness and flexibility of our method. 

\begin{figure*}[t!]
\centering
  \includegraphics[width=0.90\textwidth]{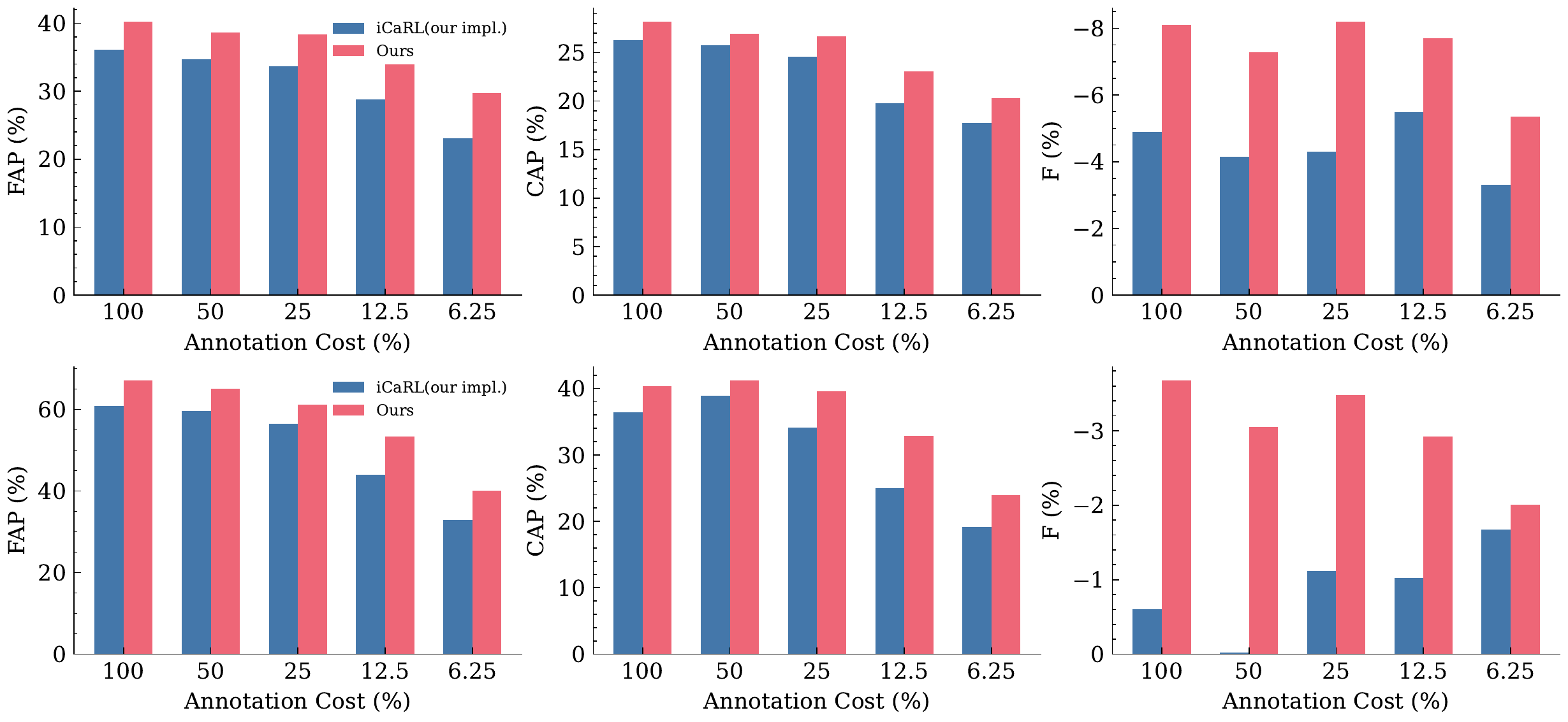}
  \caption{
  \textbf{Evaluation of online continual object detection in video streams with three metrics (FAP, CAP and F, Section~\ref{sec:eval}) on OAK dataset (first row) and EgoObjects dataset (second row)}. The higher the bars are, the better. The x-axis denotes the percentage of video frames that are labeled in the video stream. It ranges from 6.25\% to 100\% (full supervision). The y-axis indcates the performance using different evaluation metrics. Ours (iCaRL(our impl.) w/ \ours{}, red) consistently beats the comparative SOTA (iCaRL(our impl.), blue) in all evaluation metrics. 
  }
  \vspace{0.5em}
  \label{fig:anno_cost}
\end{figure*}

\subsection{Sparse Annotation Protocol} 

\paragraph{Performance of CL baselines.} The sparse annotation protocol is more challenging than the previous fully supervised protocol as shown by the performance differences when number of annotated video frames decreases (compare the performance of each colored bar along the x-axis within each subplot in Figure~\ref{fig:anno_cost}).
We noted that GDumb is more resilient against the reduce of supervision. Specifically, from Table~\ref{tab:low_anno} at the lowest annotation cost of 6.25\% on EgoObjects dataset, GDumb achieves the highest performance of 38.74\%, 22.69\% and -4.53\% in FAP, CAP and F, which surpasses other baselines by a considerable margin. Same observations can be made on OAK dataset. One possible explanation is that GDumb only trains the data stored in the balanced replay buffer, which makes it less vulnerable to class imbalance problem brought by the reduce of labeled samples in the training set.

\vspace{-0.5em}
\paragraph{Performance w/ \ours{}.} Our proposed \ours{} is a plug-and-play module that can be easily inserted into and enhance existing continual learning algorithms with the ability to use unlabeled video frames effectively. 
In both OAK and EgoObjects dataset, \ours{} consistently improves the comparative SOTAs in all three evaluation metrics regardless of various degrees of annotation cost. As shown in Table~\ref{tab:low_anno}, at a lower annotation cost of 6.25\%, \ours{} doubles the performance of DER++ and A-GEM in terms of FAP and CAP, and achieves an even larger improvement in preventing forgetting. Thanks to the useful information from pseudo labels predicted by the slow learner in \ours{}, our method is more robust to various annotation costs, compared with SOTAs (compare the rate of change of blue bars \vs{} red bars over different degrees of annotation cost). Most remarkably, \ours{} with 25\% annotation cost has already outperformed comparative SOTAs with 100\% annotation cost (see Table~\ref{tab:main}). 

\input{tables/low_anno}

\input{tables/ablation.tex}

\subsection{Ablation Study}\label{sec:ablation}
We assessed the importance of our key design choices.
The Complementary Learning Systems (CLS) in \ours{} is the key for rapidly adapting to learn new tasks, meanwhile, retaining previously learnt knowledge. It constitutes of two memory reinstatement mechanisms: one is synaptic weight transfer from fast to slow learner via exponential moving average (EMA); and the other is reciprocal replay from slow learner to fast learner with pseudo-labeling (PL). 
Here we studied their effects individually. 

\vspace{-0.5em}
\paragraph{Ablation: Exponential Moving Average (EMA).} 
We removed EMA by setting the $\alpha$ in Equation~\ref{equ:EMA} to 1, where the model weights of the fast learner and slow learner are now shared throughout the learning process. Note that in the fully supervised protocol, the pseudo-labeling is turned off and the \ours{} equals to the EMA alone. From Table~\ref{tab:main} at 100\% annotation cost, we observed that removing EMA leads to significant performance drops ranging from 2\% to 8\%, for all the comparative baselines on both OAK and EgoObjects. This shows that the slow learner can effectively consolidate the knowledge from the fast learner, and constructively alleviate catastrophic forgetting by synapses consolidation over time. 
Similar observations were made in Table~\ref{tab:ablation}  
in the sparse annotation protocol (compare Row 2 \vs{} Row 1, Row 4 \vs{} Row 3). It is worth noting that, the performance difference between naive model (Row 1) and its variant with EMA (Row 2) is slightly larger in lower supervision (\ie{} 12.5\%, 6.25\%) than higher supervision (\ie{} 50\%, 25\%).
One possible reason is that, compared with higher supervision, the fast learner suffers more forgetting in lower supervision; hence, the effect of removing EMA becomes stronger in lower supervision, again highlighting the importance of EMA.

\vspace{-0.5em}
\paragraph{Ablation: Pseudo-labeling (PL).} 
We ablate our model by removing the PL of the slow learner across varying annotation costs and reported the results in Table~\ref{tab:ablation}.
The removal of PL (Row 2) leads to a performance drop of around 2\% in FAP, 1\% in CAP and 0.5-4\% in F, compared with our full \ours{} (Row 4). It implies that the slow learner captures useful semantic information from unlabeled video frames and these predicted pseudo labels are helpful in training the fast learner. 

To investigate whether pseudo labels predicted by the fast learner itself could help stream learning, we conducted another ablation experiment where we performed PL without EMA (Row 3). Compared with the naive model (Row 1), we observed a performance increase from 28.76\% to 31.60\% in FAP and 19.80\% to 22.44\% in CAP. It indicates that, due to the temporal correlation in video stream, pseudo labels predicted by the fast learner can serve as an informative supervision for the training of the fast learner itself. 

However, replaying the self-predicted pseudo labels on the fast learner fails to prevent forgetting, as indicated by the drop from -5.48\% to -4.83\% in F. It is possible that the pseudo labels generated by the fast learner only bias towards the classes which have already been learnt very well and fail to reinforce the fast learner to improve on the poorly-learnt classes. Different from the fast learner, the slow learner integrates semantic information over time. The predicted pseudo labels carry more semantic information, which is useful for fast learner to capture more generic object representations during pseudo label replays. Again, this emphasizes that the reciprocal replay from the slow learner to the fast learner is critical for memory reinstatement, which has been missing in the computational modeling literature of CLS.

\section{Conclusion}\label{sec:conclusion}

To imitate what humans see and learn in the real world, we introduce a more realistic and challenging problem of label-efficient online continual object detection (LEOCOD) in video streams. 
Addressing this problem would greatly benefit real-world applications by reducing model retraining time and data labeling costs. 
Inspired by the Complementary Learning Systems (CLS) theory, we propose a plug-and-play module, namely \ours{}, that can be easily integrated into and improve existing continual leaning algorithms.
We evaluate \ours{} on two challenging real-world video datasets, where our method achieves the state-of-the-art performance in preventing catastrophic forgetting, all while requiring minimal annotation effort.

\paragraph{Acknowledgements}
This research is funded by the National Research Foundation, Singapore under its NRFF Award NRF-NRFF13-2021-0008 \& NRF-NRFF15-2023-0001, AI Singapore Programme (AISG Award No: AISG-GC-2019-001-2A \& AISG2-RP-2021-025), Mike Zheng Shou's Start-Up Grant from NUS, Mengmi Zhang's Startup Grant from Agency for Science, Technology, and Research (A*STAR), and Early Career Investigatorship from Center for Frontier AI Research (CFAR), A*STAR. The computational work for this article was partially performed on resources of the National Supercomputing Centre, Singapore.


{\small
\bibliographystyle{ieee_fullname}
\bibliography{main_iccv.bbl}
}

\input{appendix.tex}

\end{document}

%% file: tables/main_results.tex
\begin{table*}[t!]
\centering
\resizebox{0.98\textwidth}{!}{%
\begin{tabular}{lcccclccc}
\hline
 &
  \multicolumn{1}{l}{} &
  \multicolumn{3}{c}{OAK} &
   &
  \multicolumn{3}{c}{EgoObjects} \\ \cline{3-5} \cline{7-9} 
 &
  Annotation Cost &
  FAP ($\uparrow$) &
  CAP ($\uparrow$) &
  F ($\downarrow$) &
  \multicolumn{1}{c}{} &
  FAP ($\uparrow$) &
  CAP ($\uparrow$) &
  F ($\downarrow$) \\ \hline
Incremental &
  100\% &
  8.38 &
  7.72 &
  0.03 &
  \multicolumn{1}{c}{} &
  10.21 &
  3.55 &
  1.48 \\
Offline Training &
  100\% &
  48.28 &
  35.23 &
  - &
   &
  86.18 &
  59.81 &
  - \\ \hline
EWC &
  100\% &
  7.73 &
  7.02 &
  -0.12 &
   &
  5.15 &
  1.60 &
  0.57 \\
iOD &
  100\% &
  7.92 &
  7.14 &
  0.98 &
  \multicolumn{1}{c}{} &
  8.80 &
  2.64 &
  0.00 \\
iCaRL(Wang \etal) &
  100\% &
  22.89 &
  16.60 &
  -2.95 &
  \multicolumn{1}{c}{} &
  37.61 &
  21.71 &
  2.79 \\ \hline
iCaRL(our impl.) &
  100\% &
  36.14 &
  26.26 &
  -4.89 &
  \multicolumn{1}{c}{} &
  60.80 &
  36.41 &
  -0.60 \\
\multirow{2}{*}{\;\;w/ Efficient-CLS} &
  25\% &
  38.36(\textcolor{darkgreen}{+2.22}) &
  26.64(\textcolor{darkgreen}{+0.38}) &
  -8.20(\textcolor{darkgreen}{-3.31}) &
   &
  61.26(\textcolor{darkgreen}{+0.46}) &
  39.58(\textcolor{darkgreen}{+3.17}) &
  -3.48(\textcolor{darkgreen}{-2.88}) \\
 &
  100\% &
  40.24(\textcolor{darkergreen}{+4.10}) &
  28.18(\textcolor{darkergreen}{+1.92}) &
  -8.10(\textcolor{darkergreen}{-3.21}) &
  \multicolumn{1}{c}{} &
  67.05(\textcolor{darkergreen}{+6.25}) &
  40.36(\textcolor{darkergreen}{+3.95}) &
  -3.67(\textcolor{darkergreen}{-3.07}) \\ \hline
A-GEM &
  100\% &
  36.94 &
  26.19 &
  -5.54 &
   &
  58.79 &
  35.88 &
  -8.38 \\
\multirow{2}{*}{\;\;w/ Efficient-CLS} &
  25\% &
  37.06(\textcolor{darkgreen}{+0.12}) &
  26.36(\textcolor{darkgreen}{+0.17}) &
  -7.76(\textcolor{darkgreen}{-2.22}) &
   &
  63.06(\textcolor{darkgreen}{+4.27}) &
  39.46(\textcolor{darkgreen}{+3.58}) &
  -7.49(\textcolor{debianred}{+0.89}) \\
 &
  100\% &
  39.87(\textcolor{darkergreen}{+2.93}) &
  27.97(\textcolor{darkergreen}{+1.78}) &
  -7.17(\textcolor{darkergreen}{-1.63}) &
   &
  66.94(\textcolor{darkergreen}{+8.15}) &
  39.57(\textcolor{darkergreen}{+3.69}) &
  -11.68(\textcolor{darkergreen}{-3.30}) \\ \hline
GDumb &
  100\% &
  35.27 &
  25.29 &
  -6.59 &
   &
  58.85 &
  36.38 &
  -5.21 \\
\multirow{2}{*}{\;\;w/ Efficient-CLS} &
  25\% &
  37.67(\textcolor{darkgreen}{+2.40}) &
  25.59(\textcolor{darkgreen}{+0.30}) &
  -9.30(\textcolor{darkgreen}{-2.71}) &
   &
  62.70(\textcolor{darkgreen}{+3.85}) &
  38.78(\textcolor{darkgreen}{+2.40}) &
  -8.86(\textcolor{darkgreen}{-3.65}) \\
 &
  100\% &
  38.61(\textcolor{darkergreen}{+3.34}) &
  26.04(\textcolor{darkergreen}{+0.75}) &
  -9.14(\textcolor{darkergreen}{-2.55}) &
  \multicolumn{1}{c}{} &
  63.55(\textcolor{darkergreen}{+4.70}) &
  38.98(\textcolor{darkergreen}{+2.60}) &
  -7.50(\textcolor{darkergreen}{-2.29}) \\ \hline
DER++ &
  100\% &
  37.79 &
  25.24 &
  -2.87 &
  \multicolumn{1}{c}{} &
  55.82 &
  30.84 &
  -6.08 \\
\multirow{2}{*}{\;\;w/ Efficient-CLS} &
  25\% &
  37.93(\textcolor{darkgreen}{+0.14}) &
  25.64(\textcolor{darkgreen}{+0.4}) &
  -8.90(\textcolor{darkgreen}{-6.03}) &
   &
  59.70(\textcolor{darkgreen}{+3.88}) &
  34.15(\textcolor{darkgreen}{+3.31}) &
  -11.21(\textcolor{darkgreen}{-5.13}) \\
 &
  100\% &
  39.61(\textcolor{darkergreen}{+1.82}) &
  26.73(\textcolor{darkergreen}{+1.49}) &
  -8.30(\textcolor{darkergreen}{-5.43}) &
  \multicolumn{1}{c}{} &
  62.01(\textcolor{darkergreen}{+6.19}) &
  33.09(\textcolor{darkergreen}{+2.25}) &
  -11.05(\textcolor{darkergreen}{-4.97}) \\ \hline
\end{tabular}%
}

\vspace{2.5mm}
\caption{\textbf{Overall performance of existing algorithms and \ours{} on OAK and EgoObjects}. iCaRL(Wang \etal) denotes the SOTA model presented in \cite{wang2021wanderlust}, and iCaRL(our impl.) is the same method by our implementation.
}
\vspace{-0.5em}
\label{tab:main}

\end{table*}

%% file: tables/low_anno.tex
\begin{table}[]
\centering
\setlength{\tabcolsep}{5.5pt}
\resizebox{0.48\textwidth}{!}{%
\begin{tabular}{lccccccc}
\hline
                     & \multicolumn{3}{c}{OAK}                                &  & \multicolumn{3}{c}{EgoObjects}                         \\ \cline{2-4} \cline{6-8} 
                     & FAP ($\uparrow$) & CAP ($\uparrow$) & F ($\downarrow$) &  & FAP ($\uparrow$) & CAP ($\uparrow$) & F ($\downarrow$) \\ \hline
iCaRL                & 23.04            & 17.75            & -3.31            &  & 32.91            & 19.15            & -1.36            \\
\;\;w/ Efficient-CLS & \textbf{29.72}   & \textbf{20.31}   & \textbf{-5.36}   &  & \textbf{40.16}   & \textbf{23.95}   & \textbf{-2.01}   \\ \hline
A-GEM                & 23.59            & 16.15            & -2.99            &  & 21.84            & 12.28            & 0.82             \\
\;\;w/ Efficient-CLS & \textbf{30.18}   & \textbf{20.59}   & \textbf{-5.44}   &  & \textbf{38.96}   & \textbf{23.79}   & \textbf{-5.64}   \\ \hline
GDumb                & 27.37            & 19.64            & -4.25            &  & 38.74            & 22.69            & -4.53            \\
\;\;w/ Efficient-CLS & \textbf{29.07}   & \textbf{19.99}   & \textbf{-6.01}   &  & \textbf{40.09}   & \textbf{23.67}   & \textbf{-5.16}   \\ \hline
DER++                & 24.21            & 15.93            & -3.79            &  & 16.95            & 8.48             & 2.03             \\
\;\;w/ Efficient-CLS & \textbf{28.63}   & \textbf{19.64}   & \textbf{-4.60}   &  & \textbf{35.78}   & \textbf{20.74}   & \textbf{-4.69}   \\ \hline
\end{tabular}
}
\vspace{1mm}
\caption{\textbf{Performance of \ours{} at low annotation cost of 6.25\%}. The best results are \textbf{bold-faced}.}
\label{tab:low_anno}
\vspace{-0.5em}
\end{table}

%% file: tables/ablation.tex
\begin{table*}[h!]
\centering
\resizebox{0.96\textwidth}{!}{%
\begin{tabular}{ccccccccccccccccc}
\hline
       &        & \multicolumn{3}{c}{50\%} &  & \multicolumn{3}{c}{25\%} &  & \multicolumn{3}{c}{12.5\%} &  & \multicolumn{3}{c}{6.25\%} \\ \cline{3-5} \cline{7-9} \cline{11-13} \cline{15-17} 
EMA &
  PL &
  FAP ($\uparrow$) &
  CAP ($\uparrow$) &
  F ($\downarrow$) &
   &
  FAP ($\uparrow$) &
  CAP ($\uparrow$) &
  F ($\downarrow$) &
   &
  FAP ($\uparrow$) &
  CAP ($\uparrow$) &
  F ($\downarrow$) &
   &
  FAP ($\uparrow$) &
  CAP ($\uparrow$) &
  F ($\downarrow$) \\ \hline
\xmark & \xmark & 34.68  & 25.78  & -4.15  &  & 33.70  & 24.57  & -4.30  &  & 28.76   & 19.80   & -5.48  &  & 23.04   & 17.75   & -3.31  \\
\cmark & \xmark & 35.74  & 25.77  & -4.82  &  & 34.79  & 25.62  & -4.35  &  & 31.72   & 21.16   & -7.24  &  & 27.84   & 20.03   & -3.96  \\
\xmark & \cmark & 35.61  & 25.56  & -3.76  &  & 34.95  & 25.65  & -3.65  &  & 31.60   & 22.44   & -4.83  &  & 26.39   & 19.50   & -1.99  \\
\cmark &
  \cmark &
  \textbf{38.61} &
  \textbf{26.90} &
  \textbf{-7.29} &
   &
  \textbf{38.36} &
  \textbf{26.64} &
  \textbf{-8.20} &
   &
  \textbf{33.92} &
  \textbf{23.04} &
  \textbf{-7.71} &
   &
  \textbf{29.72} &
  \textbf{20.31} &
  \textbf{-5.36} \\ \hline 
\end{tabular}%
}
\vspace{2mm}
\caption{\textbf{Effectiveness of Exponential Moving Average (EMA) and Pseudo-labeling (PL) for iCaRL(our impl.) on OAK dataset at annotation cost 50\%, 25\%, 12.5\% and 6.25\%}. The best results are \textbf{bold-faced}.}
\label{tab:ablation}
\vspace{-0.5em}
\end{table*}

%% file: appendix.tex
\newpage
\appendix
\appendixpage

\setcounter{table}{0}
\setcounter{figure}{0}
\setcounter{equation}{0}
\renewcommand{\thetable}{S\arabic{table}}
\renewcommand{\theHtable}{S\arabic{table}}
\renewcommand{\thefigure}{S\arabic{figure}}
\renewcommand{\theHfigure}{S\arabic{figure}}
\renewcommand{\theequation}{S\arabic{equation}}
\renewcommand{\theHequation}{S\arabic{equation}}

{
\hypersetup{linkcolor=black}
\addtocontents{toc}{\cftpagenumbersoff{section}} 
\setlength{\cftbeforesecskip}{6pt}
\startcontents[Table of Content]
\printcontents[Table of Content]{l}{1}{\section*{Table of Content}\setcounter{tocdepth}{2}}
}

\section{Additional Results}

\subsection{Effect of Pseudo-labeling Threshold}

As mentioned in Section~\ref{sec:method},
we apply confidence thresholding to remove predicted bounding boxes that have low confidence scores. To show the effectiveness of thresholding, we vary the confidence threshold $\tau$ from 0.1 to 0.9 (see Table~\ref{stab:pseudo_thresh}). We observed that the model using a high threshold (\eg{}, 0.7) yields satisfactory results, as it produces more reliable pseudo-labels with high confidence. On the other hand, using a low threshold can result in lower performance since the model generates too many bounding boxes, which are likely to be false positives.

\input{tables/suppl_pseudo_thresh.tex}

\subsection{Effect of RPN Loss in Pseudo Training}\label{subsec:rpn_loss}

In our \ours{}, pseudo losses are only applied at the ROI module but not at the RPN module. As shown in Table~\ref{stab:rpn_loss}, the model with and without RPN loss in training pseudo-labeled frames show similar performance. We conjecture that the RPN module is less likely to suffer catastrophic forgetting since its primary function is to produce general proposals that are class agnostic. As a result, we removed the RPN loss during pseudo training, which also reduces the overall computational cost.

\input{tables/suppl_rpn_loss.tex}

\subsection{Effect of Pseudo Loss Weights}

As mentioned in Section~\ref{sec:method}, 
$\lambda_{pseudo}$ is a hyperparameter balancing the importance of supervised loss ($\mathcal{L}_{sup}$) and pseudo loss ($\mathcal{L}_{pseudo}$). To examine the effect of $\lambda_{pseudo}$, we vary the $\lambda_{pseudo}$ from 0.5 to 4.0 at annotation cost 12.5\% on OAK dataset. As shown in Table~\ref{stab:lambda_pseudo}, the model performs the best with $\lambda_{pseudo}=1.0$ and shows moderate performance drop for other values of $\lambda_{pseudo}$ (0.5, 1.5, and 2.0). However, when $\lambda_{pseudo}$ is set to 4.0, the model performance deteriorates.

\input{tables/suppl_pseudo_lambda.tex}

\subsection{Effect of Data Augmentation}

We use data augmentation techniques when training the pseudo-labeled frames. Here we ablate our \ours{} by removing data augmentation in pseudo training. From Table~\ref{stab:aug} at annotation cost 12.5\%, we observed that removing data augmentation in pseudo training leads to a performance drop of 3.60\% in FAP, 1.96\% in CAP and 1.21\% in F on OAK dataset. This indicates that using data augmentation on pseudo-labeled frames can enforce the model to learn invariant object representations from these video frames.

\input{tables/suppl_data_aug.tex}

\subsection{Design Decisions on External Replay Buffer}\label{subsec:buffer_design}

First, \textbf{we explore whether a replay buffer needs to be balanced based on class distribution in the video streams}. Our current design ensures that there are at least 5 images containing object instances of any given learnt classes. In comparison, we conducted an additional experiment (Random Store and Replay) where we designed a replay buffer of the same size as \ours{} and saved any new images regardless of the class labels. Ideally, this replay buffer represents the imbalanced class distribution of the video streams. For example, cars appear more often than stop signs; thus, it is more likely to store more images containing cars in the replay buffer. In Table~\ref{stab:balanced_replay}, we found that a class balanced replay buffer performs much better than the one that randomly stores and replays, implying the importance of class balanced design for replay buffer. 

\input{tables/suppl_balanced_replay.tex}

Second, \textbf{we study how many frames are needed to retrieve from the external buffer for replay} in conjunction with the batch of video frames in the current training iteration (Table~\ref{stab:replay_size}). To perform episodic replay, we randomly retrieve 16 video frames from the replay buffer for joint training with current frames of a mini-batch. As shown in Table~\ref{stab:replay_size}, replaying 16 video frames is a good trade-off between model performance and extra training time. Replaying fewer samples would lead to poor performance and forgetting, while replaying more hardly brings any benefits but largely increases the training resources. We compare our method with the implementation of Wang et al.~\cite{wang2021wanderlust} where the batch size for replays increases according to the number of seen class in each iteration. Table~\ref{tab:main} demonstrates the superior performance of our replay technique.

\input{tables/suppl_replay_size.tex}

Third, \textbf{we study the effect of the number of sample image stored per class in the replay buffer}. We vary the number of sample image saved per class. As the number of sample image per class increases, it increases the diversity of the object representation per class; hence leads to steady performance boost (Table~\ref{stab:buffer_size}). This trend is observed in both iCaRL and \ours{}. Of course, one could argue that the ideal case is to store all the past video frames and replay all of them. This reverts to the offline setting and yields the best model performance in LEOCOD. However, it is at the expense of heavy usage of memory storage. In practice, one has to strike a nice balance between model performance and memory storage. Here, we demonstrate that even saving 5 images per class, \ours{} surpasses all the competitive baselines in LEOCOD.

\input{tables/suppl_buffer_size.tex}

\subsection{Analysis of Unlabeled Frames Selection}\label{subsec:unlabeled_sel}

We conduct experiments with 5 runs. Each run applies a different random seed for unlabeled frames selection. We report the means and standard deviations of these 5 runs in Table~\ref{stab:seed}. We found that our \ours{} shows reliable and robust performance against different selections of unlabeled frames in the video stream.

\input{tables/suppl_seed.tex}

\begin{figure*}[h!]%
\centering
\begin{center}
\begin{tabular}{c}
\includegraphics[scale=0.55]{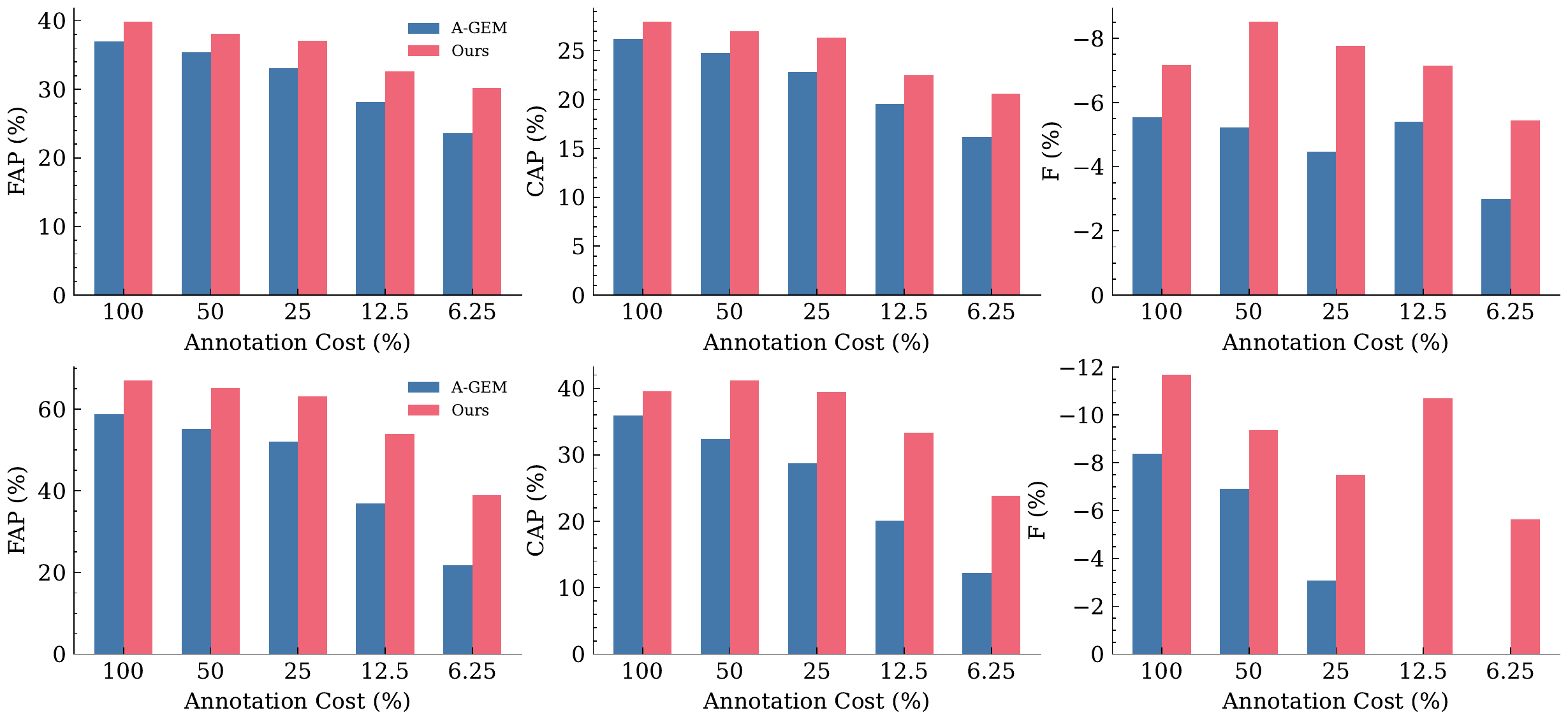}\\
A) Comparisons of A-GEM (blue) and A-GEM \emph{w/} \ours{} (red).\\[3ex]
\includegraphics[scale=0.55]{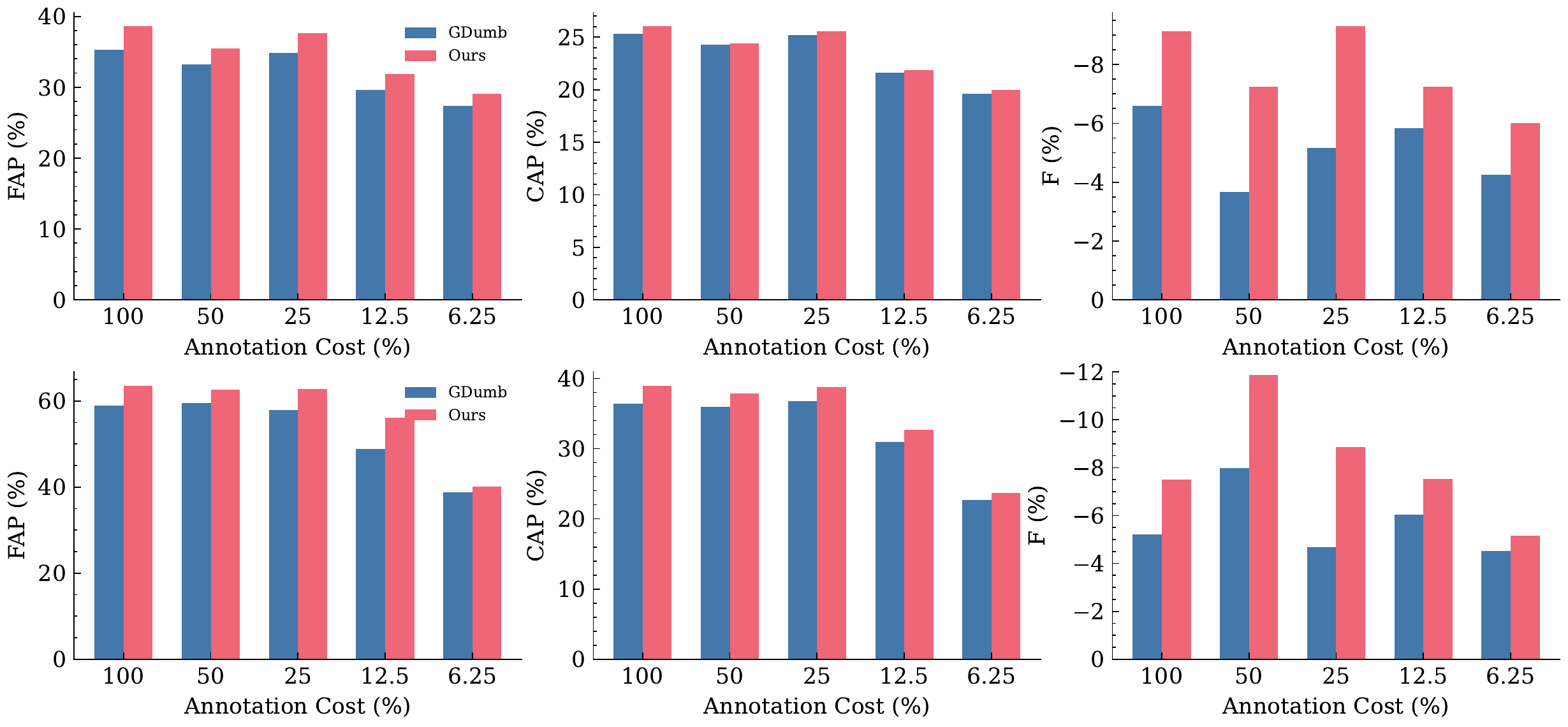}\\
B) Comparisons of GDumb (blue) and GDumb \emph{w/} \ours{} (red).\\[3ex]
\includegraphics[scale=0.55]{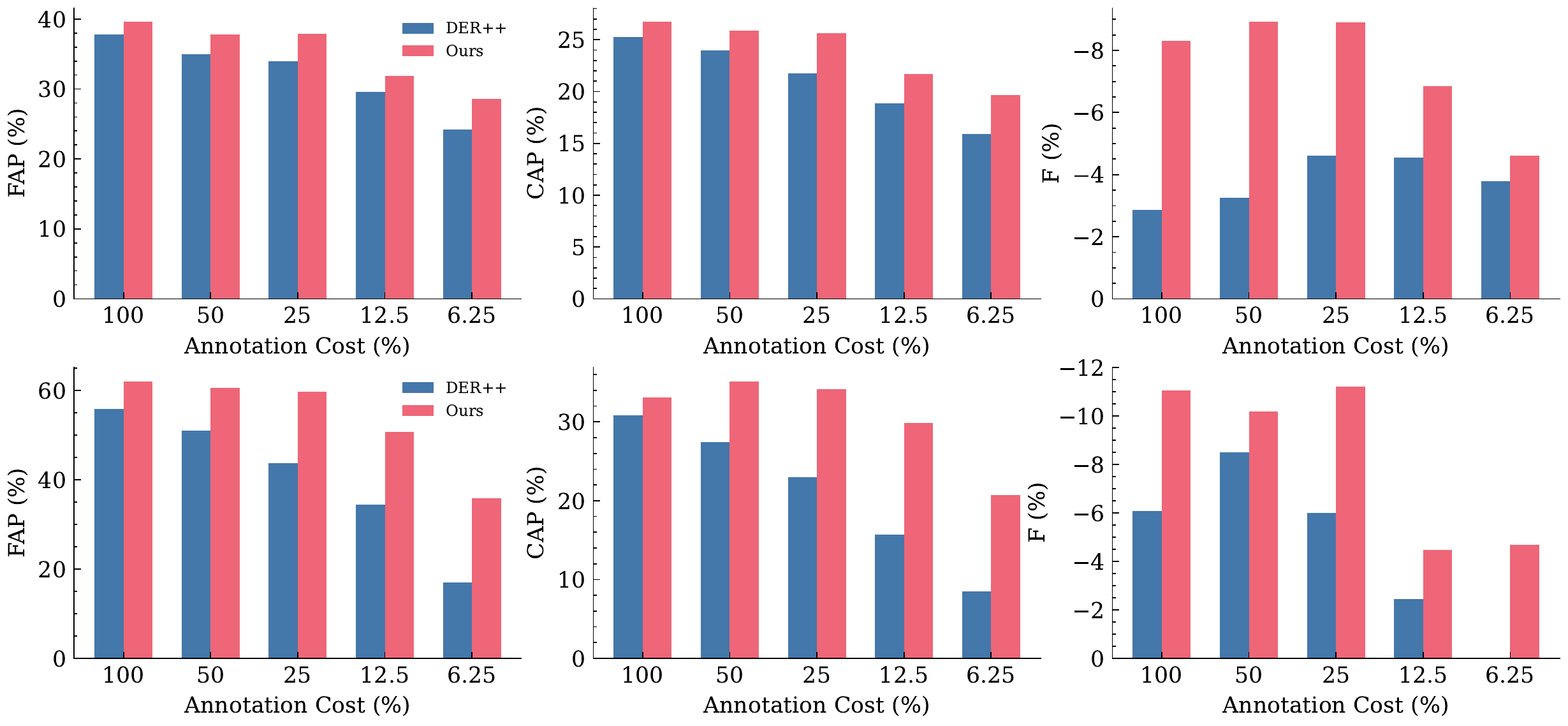}\\
C) Comparisons of DER++ (blue) and DER++ \emph{w/} \ours{} (red).
\end{tabular}
\end{center}
\caption{\textbf{Evaluation of state-of-the-art CL methods in LEOCOD setting on OAK dataset (first row) and EgoObjects dataset (second row)}. The higher the bars are, the better. The x-axis denotes the percentage of video frames that are labeled in the video stream. It ranges from 6.25\% to 100\% (full supervision). The y-axis indcates the performance using different evaluation metrics.}%
\label{sfig:sparse}%
\end{figure*}

\subsection{Visualization of Pseudo-labeling}

\begin{figure*}[h!]
\centering
  \includegraphics[width=0.96\textwidth]{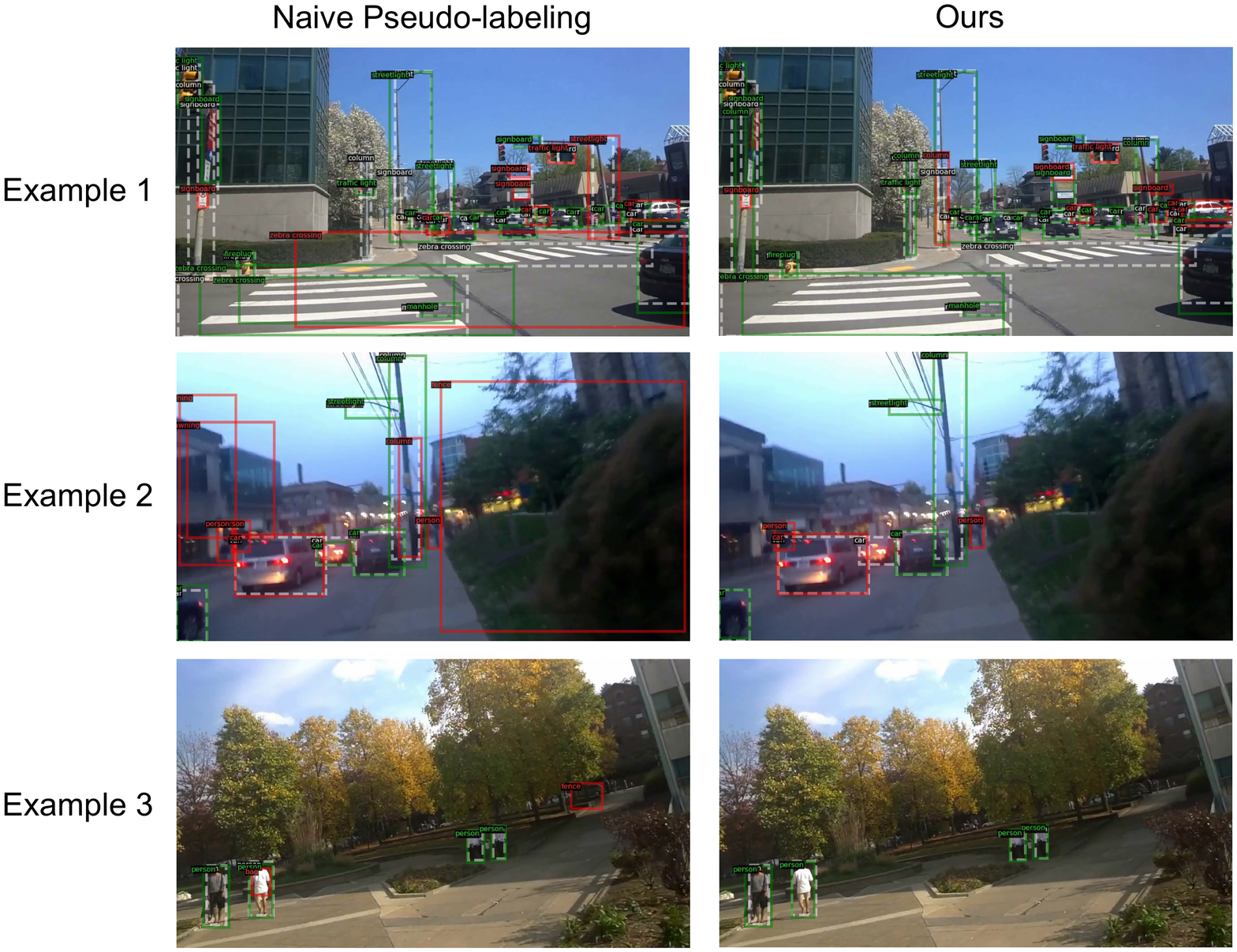}
  \caption{\textbf{Visualization of example pseudo-labels predicted by our \ours{} and the Naive Pseudo-labeling.}
  The white box with dash line denotes the ground truth label. The box with solid line denotes the pseudo-labels (the ones in green are correct while the red are wrong labels). The \textbf{Naive Pseudo-labeling} only has one learner and uses the pseudo-labels generated by itself for training. }
  \vspace{-1em}
  \label{sfig:vis}
\end{figure*}

As shown in Figure~\ref{sfig:vis}, the pseudo-labels generated by our \ours{} (2nd column) capture more ground truth objects and contain fewer false positive instances than the Naive Pseudo-labeling model (1st column).

\section{Limitations and Future Work.} 
Same as other replay methods, our method is facing with infinite memory expansion problem. Our replay buffer stores 5 images per object class. As the number of seen class increases, the memory buffer has to expand. One could imagine that this will be a challenging case in lifelong learning on video streams which last for tens of years and thousands of classes have to be learnt. In the future, we will explore more efficient replay strategies with latent object representations. 

%% file: tables/suppl_pseudo_thresh.tex
\begin{table}[h!]
\centering
\resizebox{0.48\textwidth}{!}{%
\begin{tabular}{cccccccc}
\hline
$\tau$           & 0.1   & 0.3   & 0.5   & 0.6   & 0.7   & 0.8   & 0.9   \\ \hline
FAP ($\uparrow$) & 30.33 & 31.16 & 32.17 & 32.54 & 33.92 & 32.81 & 32.07 \\
CAP ($\uparrow$) & 21.45 & 21.67 & 22.38 & 22.51 & 23.04 & 22.69 & 22.70 \\
F ($\downarrow$) & -7.11 & -7.29 & -6.88 & -6.98 & -7.71 & -6.60 & -6.94 \\ \hline
\end{tabular}
}
\vspace{1mm}
\caption{\textbf{Ablation study of varying confidence threshold $\tau$ on OAK dataset at annotation cost 12.5\%.}}
\label{stab:pseudo_thresh}
\end{table}

%% file: tables/suppl_rpn_loss.tex
\begin{table}[h!]
\centering
\begin{tabular}{cccc}
\hline
RPN Loss & FAP ($\uparrow$) & CAP ($\uparrow$) & F ($\downarrow$) \\ \hline
\xmark   & \textbf{33.92}   & \textbf{23.04}   & \textbf{-7.71}   \\
\cmark   & 33.64            & 22.68            & -7.62            \\ \hline
\end{tabular}
\vspace{3mm}
\caption{\textbf{Effect of RPN loss in pseudo training on OAK dataset at annotation cost 12.5\%}.}
\label{stab:rpn_loss}
\end{table}

%% file: tables/suppl_pseudo_lambda.tex
\begin{table}[h!]
\centering
\begin{tabular}{cccccc}
\hline
$\lambda_{pseudo}$ & 0.5   & 1.0   & 1.5   & 2.0   & 4.0   \\ \hline
FAP ($\uparrow$) & 33.19 & 33.92 & 33.01 & 32.67 & 30.08 \\
CAP ($\uparrow$) & 22.92 & 23.04 & 22.52 & 22.02 & 20.17 \\
F ($\downarrow$) & -6.55 & -7.71 & -7.71 & -7.86 & -7.50 \\ \hline
\end{tabular}
\vspace{3mm}
\caption{\textbf{Ablation study of varying pseudo loss weights $\lambda_{pseudo}$ on OAK dataset at annotation cost 12.5\%.}}
\label{stab:lambda_pseudo}
\vspace{-2mm}
\end{table}

%% file: tables/suppl_data_aug.tex
\begin{table}[h!]
\centering
\begin{tabular}{cccc}
\hline
Data Augmentation & FAP ($\uparrow$) & CAP ($\uparrow$) & F ($\downarrow$) \\ \hline
\xmark            & 30.32            & 21.08            & -6.50            \\
\cmark            & \textbf{33.92}   & \textbf{23.04}   & \textbf{-7.71}   \\ \hline
\end{tabular}
\vspace{3mm}
\caption{\textbf{Effect of data augmentation in pseudo training on OAK dataset at annotation cost 12.5\%}.}
\vspace{-3mm}
\label{stab:aug}
\end{table}

%% file: tables/suppl_balanced_replay.tex
\begin{table}[h!]
\centering
\resizebox{0.48\textwidth}{!}{%
\begin{tabular}{cccc}
\hline
 & FAP ($\uparrow$) & CAP ($\uparrow$) & F ($\downarrow$) \\ \hline
Random Store and Replay & 26.23 & 21.45 & -3.23 \\
Balanced Store and Replay & \textbf{40.24} & \textbf{28.18} & \textbf{-8.10} \\ \hline
\end{tabular}
}
\vspace{1mm}
\caption{\textbf{Effect of class balanced replay buffer on OAK dataset in fully supervised protocol.}}
\label{stab:balanced_replay}
\vspace{-4mm}
\end{table}

%% file: tables/suppl_replay_size.tex
\begin{table}[th!]
\centering
\resizebox{0.48\textwidth}{!}{%
\begin{tabular}{ccccc}
\hline
 & \multicolumn{1}{l}{Replay Size} & FAP ($\uparrow$) & CAP ($\uparrow$) & F ($\downarrow$) \\ \hline
\multirow{6}{*}{iCaRL (our impl.)} & 2 & 14.14 & 10.62 & -0.09 \\
 & 4 & 21.93 & 15.26 & -0.57 \\
 & 8 & 34.08 & 22.20 & -4.83 \\
 & \textbf{16} & \textbf{36.14} & \textbf{26.26} & \textbf{-4.89} \\
 & 32 & 37.42 & 28.28 & -3.09 \\
 & 64 & 35.59 & 27.81 & -1.65 \\ \hline
\multirow{6}{*}{Efficient-CLS} & 2 & 18.79 & 13.37 & -2.05 \\
 & 4 & 25.52 & 18.22 & -4.64 \\
 & 8 & 37.03 & 24.32 & -8.21 \\
 & \textbf{16} & \textbf{40.24} & \textbf{28.18} & \textbf{-8.10} \\
 & 32 & 41.04 & 30.01 & -7.90 \\
 & 64 & 38.79 & 29.92 & -7.61 \\ \hline
\end{tabular}
}
\vspace{2mm}
\caption{\textbf{Ablation study of the number of replay sample per training step on OAK dataset in fully supervised protocol.} The buffer size is set to 5 images per class. Our choices are \textbf{bold-faced}.}
\label{stab:replay_size}
\vspace{-1mm}
\end{table}

%% file: tables/suppl_buffer_size.tex
\begin{table}[th!]
\centering
\resizebox{0.48\textwidth}{!}{%
\begin{tabular}{ccccc}
\hline
 & \multicolumn{1}{l}{Buffer Size} & FAP ($\uparrow$) & CAP ($\uparrow$) & F ($\downarrow$) \\ \hline
\multirow{7}{*}{iCaRL (our impl.)} & 1 & 28.60 & 20.60 & -2.59 \\
 & \textbf{5} & \textbf{36.14} & \textbf{26.26} & \textbf{-4.89} \\
 & 10 & 39.96 & 28.37 & -6.45 \\
 & 15 & 40.56 & 28.84 & -6.69 \\
 & 20 & 41.26 & 29.16 & -7.19 \\
 & 30 & 42.04 & 29.57 & -7.57 \\
 & 50 & 43.22 & 30.21 & -7.53 \\ \hline
\multirow{7}{*}{Efficient-CLS} & 1 & 31.41 & 22.50 & -6.74 \\
 & \textbf{5} & \textbf{40.24} & \textbf{28.18} & \textbf{-8.10} \\
 & 10 & 43.44 & 29.96 & -8.80 \\
 & 15 & 44.57 & 30.17 & -8.98 \\
 & 20 & 45.10 & 30.36 & -8.46 \\
 & 30 & 45.26 & 30.87 & -9.37 \\
 & 50 & 46.95 & 31.33 & -9.00 \\ \hline
\end{tabular}
}
\vspace{2mm}
\caption{\textbf{Ablation study of varying numbers of samples per class stored in replay buffer on OAK dataset in fully supervised protocol.} The replay size is set to 16 images per training iteration. Our choices are \textbf{bold-faced}.}
\label{stab:buffer_size}
\vspace{-2mm}
\end{table}

%% file: tables/suppl_seed.tex
\begin{table}[h!]
\centering
\resizebox{0.48\textwidth}{!}{%
\begin{tabular}{clllc}
\hline
\multicolumn{1}{l}{Annotation Cost (\%)} &
  \multicolumn{1}{c}{50} &
  \multicolumn{1}{c}{25} &
  \multicolumn{1}{c}{12.5} &
  6.25 \\ \hline
FAP ($\uparrow$) &
  \multicolumn{1}{c}{38.45 ($\pm$0.68)} &
  \multicolumn{1}{c}{38.00 ($\pm$1.17)} &
  \multicolumn{1}{c}{34.29 ($\pm$0.76)} &
  30.50 ($\pm$1.07) \\
CAP ($\uparrow$) &
  26.85 ($\pm$0.24) &
  26.38 ($\pm$0.32) &
  23.47 ($\pm$0.73) &
  \multicolumn{1}{l}{20.60 ($\pm$0.43)} \\
F ($\downarrow$) &
  -8.01 ($\pm$0.77) &
  -8.32 ($\pm$0.98) &
  -7.30 ($\pm$0.82) &
  -6.28 ($\pm$0.69) \\ \hline
\end{tabular}
}
\vspace{2mm}
\caption{\textbf{Performance of our \ours{} on OAK dataset in sparse annotation protocol.} The table header denotes the percentage of frames that are labeled in the video stream. The means and standard deviations in brackets are reported.
}
\label{stab:seed}
\vspace{-2mm}
\end{table}